%% file: main.tex
\pdfoutput=1
\documentclass[11pt]{article}

\usepackage{ACL2023}
\usepackage{times}
\usepackage{latexsym}

\usepackage[T1]{fontenc}
\usepackage[utf8]{inputenc}

\usepackage{microtype}

\usepackage{inconsolata}

\usepackage{hyperref}       % hyperlinks
\usepackage{url}            % simple URL typesetting
\usepackage{booktabs}       % professional-quality tables
\usepackage{amsfonts}       % blackboard math symbols
\usepackage{nicefrac}       % compact symbols for 1/2, etc.
\usepackage{microtype}      % microtypography
\usepackage{xcolor}         % colors
\usepackage{amsmath}
\usepackage{caption}
\usepackage{subcaption}
\usepackage{wrapfig}
\usepackage{graphicx}
\usepackage{tabularx}
\usepackage{arydshln}
\usepackage{algorithm}
\usepackage{algorithmic}

\DeclareMathOperator*{\argmax}{arg\,max}

\usepackage{amsmath}
\usepackage{amssymb}
\usepackage{mathtools}
\usepackage{amsthm}
\usepackage{bm}
\usepackage{multirow}
\usepackage{stmaryrd}
\usepackage{bbm}
\usepackage{longtable}
\usepackage{mdframed}

\usepackage{xspace}
\usepackage{paralist}
\usepackage{booktabs} 
\usepackage{ulem}
\definecolor{teaserblue}{RGB}{242, 242, 255}

\newcommand{\HOne}{$\mathcal{H}_1$} 
\newcommand{\HTwo}{$\mathcal{H}_2$}

\usepackage{adjustbox}

\title{Unraveling the Mechanics of Learning-Based Demonstration Selection for In-Context Learning}

\author{%
Hui Liu\textsuperscript{1}, Wenya Wang\textsuperscript{2}, Hao Sun\textsuperscript{3}, Chris Xing Tian\textsuperscript{1},\\
\textbf{Chenqi Kong\textsuperscript{2}, Xin Dong\textsuperscript{4}, Haoliang Li\textsuperscript{1}} \\
\textsuperscript{1}City University of Hong Kong, Hong Kong\quad \textsuperscript{2}Nanyang Technological University, Singapore \\
\textsuperscript{3}Peking University, Beijing, China \quad \textsuperscript{4}NVIDIA, America \\
\texttt{\{liuhui3-c, xingtian4-c, haoliang.li\}@my.cityu.edu.hk} \\
\texttt{\{wangwy, chenqi.kong\}@ntu.edu.sg} \\
\texttt{sunhao@stu.pku.edu.cn, xind@nvidia.com} \\
}

\begin{document}

\maketitle

\begin{abstract}
\input{0.abstract}
\end{abstract}

\input{1.intro}

\input{2.related}
\input{3.assumption}
\input{4.method}
\input{5.exp}

\input{7.conclusion}

\bibliographystyle{acl_natbib}
\bibliography{ref.bib}
\newpage
\input{6.appendix}
%%%%%%%%%%%%%%%%%%%%%%%%%%%%%%%%%%%%%%%%%%%%%%%%%%%%%%%%%%%%

\end{document}

%% file: 0.abstract.tex
Large Language Models (LLMs) have demonstrated impressive in-context learning (ICL) capabilities from few-shot demonstration exemplars. Recent learning-based demonstration selection methods have proven beneficial to ICL by choosing more useful exemplars. While these methods generally assume they learn better similarity measurements between exemplars and test cases, what kinds of similarities are captured by them and are vital to performing ICL still remain under-explored. To dive into this question, we analyze the working mechanism of learning-based demonstration selection methods and empirically identify two essential factors of their similarity measurements: 1) Integrating task-agnostic similarities of different levels between the input of exemplars and test cases; 2) Incorporating task-specific similarity between the output of exemplars and test cases. We validate these two findings through extensive quantitative analysis across ten datasets and various LLMs. Based on these insights, we introduce two simplified exemplar selection methods, MLSM and TTF, catering to task-agnostic and task-specific demands to eliminate costly data collection. The effectiveness of both methods supports our findings and pave the way for future studies.
 

%% file: 1.intro.tex
\section{Introduction}
In-context learning (ICL) has emerged as a promising paradigm that employs a sequence of demonstration exemplars as prompts to assist large language models (LLMs) in effectively performing unseen tasks~\cite{nips20llmfewshotlearner, votek}. However, the performance of ICL can be sensitive to the choice, format, and order of the in-context exemplar~\cite{ICLfactor1, ICLfactor2, ICLformat1, ACL23order2}. To mitigate this challenge, given a test case $x^t$, the exemplar selection task assumes access to a demonstration set $\mathcal{D}$ containing input-output pairs $(x,y)$ and focuses on selecting the most effective exemplar from $\mathcal{D}$ to inform the target output $y^t$.

To address this task, it is the most common practice to select demonstration exemplars based on a similarity measurement between $x$ and $x^t$~\cite{epr, CEIL, emnlp23skillselection1, diversity1, emnlp23skillselection3, ACL23order3}. Some work utilizes task-agnostic similarity like term frequency-based similarity BM25 and semantic similarity computed by off-the-shelf text encoders~\cite{ACL22retriever1, emnlp23skillselection1}. Recent learning-based  studies~\cite{epr, CEIL, li-etal-2023-unified}, however, separately train a retriever to learn implicit similarity measurements using a contrastive leaning-based proxy task where positive exemplars $x^+$ and negative exemplars $x^-$ are labeled by interacting with LLMs. This data creation process often requires hundreds of thousands of queries to LLMs for each task to collect sufficient positive/negative data. 

Although learning-based methods consistently exhibit significant performance improvements over task-agnostic similarity across various tasks, the implicit similarity they capture and their connection to the performance of ICL remain unclear. Through a detailed examination of previous works, we observe 1)  While the low-level similarity like BM25 and semantic similarity excel in different tasks (e.g., Top-K BM25 outperforms Top-K BERT on Nl2Bash~\cite{datasetnl2bash} and SWAG~\cite{dataset:swag} in Table~\ref{man:cls} and Table~\ref{man:gen}), learning-based similarity generally performs well across all tasks. 2) In the proxy task, the input and output similarity between positive exemplars and test cases is higher than that of negative exemplars and test cases. Moreover, learning-based methods often suffer from poor generalization across different tasks, as corroborated by findings in~\cite{CEIL}. Based on these initial observations, we propose two hypotheses regarding learning-based methods:

\HOne: After training, the retriever acts as an ensemble model that adaptively integrates multi-level task-agnostic similarities between the exemplar input ($x$) and test cases ($x^t$) for different tasks.

\HTwo: Beyond input similarities, the training process encourages selecting exemplars with similar output ($y$) to the output of the test case ($y^t$), implicitly predicted during retrieval, enhancing the retriever's discriminative power for a specific task.

Extensive quantitative experiments are designed to validate these hypotheses: 1) We take various layers of BERT as anchors for similarities of different levels and discover learning-based methods exhibiting varying preferences for these anchors before and after training,  suggesting an adaptive combination of these similarities tailored to different tasks. 2) We investigate the exemplar retrieved by learning-based methods and find these exemplars show a higher similarity in output to the test case than other task-agnostic similarity-based methods. This finding indicates that learning-based methods incorporate task-specific similarities between the outputs of exemplars and test cases during the exemplar selection process, potentially capturing the joint distribution of inputs and outputs between exemplars and test cases. Additionally, by connecting our findings with existing interpretative theories of ICL~\cite{iclinductionhead, randomlabeliswrong1, Repetitionsiclr24, halawi2023overthinking, anchors}, we further qualitatively validate our conclusions.

Drawing insights from these findings, we propose two cost-effective exemplar selection methods: 1) \textit{Multi-level Similarity Maximization} (MLSM) retriever that maximizes agreement across different similarity levels represented by various layers of BERT in the inference of LLMs. 2) \textit{Test Task Fine-tuning} (TTF) retriever, which uses labeled data from the demonstration set to finetune the retriever to learn task-specific information. Both retrievers eliminate the need for costly data collection for the proxy task, catering to cross-task and task-specific demands. To validate the effectiveness of these methods, we conduct experiments across five distinct LLMs and a range of tasks. These promising applications confirm our hypotheses and benefit future demonstration selection studies for more efficient LLM deployment.

%% file: 2.related.tex
\section{Preliminary}
\label{sec:preliminary}
\subsection{Learning-based Demonstration Selection}\label{sec:2.1}
Demonstration selection aims to identify a sequence of high-quality exemplars from the demonstration set as a prompt to enhance test case accuracy on LLMs. Prior studies  \cite{ACL22retriever1, DBLP:conf/acl/GaoFC20} find that good exemplars exhibit similarities with the test case. They employ the pre-trained text encoder like BERT \cite{bert} as a retriever to encode inputs and take the average embedding of all tokens from the final layer of this encoder to represent test cases and exemplars. Subsequently, cosine similarity scores are computed between test cases and exemplars to retrieve the top-K most similar exemplars as prompts. 

While the pipeline of learning-based demonstration selection methods~\cite{epr, CEIL, li-etal-2023-unified} is similar to the above strategy, they further exploit LLMs to label positive and negative exemplars to construct a proxy task to fine-tune the retriever, aiming to learn a better similarity metric. Specifically, let $\mathcal{D}$ denote the demonstration set. Given an exemplar $(x_i, y_i)$ in $\mathcal{D}$, \citet{epr} propose EPR to sample a sequence of candidate examples from $\mathcal{D}$, denoted as $S = \{(\overline{x}_1, \overline{y}_1), \ldots, (\overline{x}_m, \overline{y}_m)\}$ and score them by $s(\overline{x}, \overline{y})=P_{\mathrm{LLM}}(Y=y_i|(\overline{x}, \overline{y}), x_i)$, corresponding to the probability of producing correct output $y_i$ for $x_i$ conditioned on $(\overline{x}, \overline{y})$ using an LLM. Subsequently, $\overline{x}$ with the highest score is selected as the positive sample, denoted as $x^{+}$  and the lowest as the hard negative sample, denoted as $x^{-}$ for $x_i$. These samples are then used to train the retriever by maximizing the similarity between $x$ and $x^{+}$ and minimizing the similarity between $x$ and $x^{-}$ via contrastive learning.  In subsequent sections, without special note, we analyze EPR to unravel the mechanics of learning-based demonstration selection methods and adopt BERT\footnote{\url{https://huggingface.co/google-bert/bert-base-uncased}} consisting of twelve transformer layers as the retriever.

\subsection{Layers of BERT as Anchors of Multi-level Similarities}\label{sec:2.2}
Previous studies \cite{bertembeddingana1, bertembeddingana2, jawahar2019does} have empirically shown that the intermediate layers of BERT encode a rich hierarchy of linguistic information with surface features at the bottom, syntactic features in the middle and semantic features at the top through probing tasks. Moreover, BERT has been pre-trained on a vast corpus capturing general linguistic features that can be utilized for various tasks. These inspire us to take different layers of the original BERT (i.e., BERT without task-specific fine-tuning) as anchors of multi-level similarities. Especially for a layer $l$, given two texts $s_1$ and $s_2$, we can extract all token embedding from this layer and compute their average pooling as the representation of both texts, denoted as $h^l_1$ and $h^l_2$. Then, the similarity between $s_1$ and $s_2$ corresponding to layer $l$ can be obtained by computing the cosine similarity between $h^l_1$ and $h^l_2$.

%% file: 3.assumption.tex
\section{Rethinking Learning-based Demonstration Selection}
\label{sec:assumption}
This section proposes two key hypotheses regarding the underlying similarity mechanism of learning-based exemplar selection methods: (\HOne) The learning-based retriever is analogous to an ensemble model which adaptively aggregates multi-level similarities computed by different BERT layers between the input of exemplar and test cases ($x$ and $x^t$). (\HTwo) The learning-based retriever favors selecting exemplars with similar output ($y$) to the test case output ($y^t$). Both hypotheses are validated through quantitative analysis as shown below and qualitative analysis in Appendix~\ref{app:icl}.
% Both hypotheses are validated through quantitative analysis as below and qualitative analyses in Appendix~\ref{app:icl}. quantitative validation is conducted for both hypotheses.

\subsection{Multi-level Similarity (\HOne)}
\label{sec:HierarchicalAnalogs}
\begin{figure*}[tb]
    \centering
    \vspace{-4ex}
        \centering
        \includegraphics[width=\textwidth]{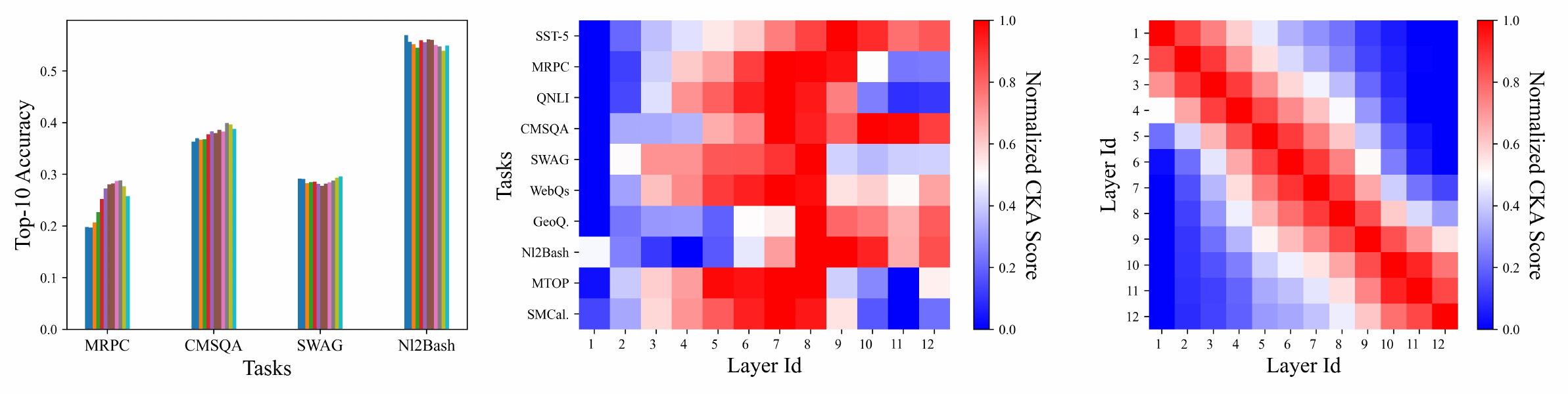}
           \vspace{-4ex}
    \caption{\textbf{Left}: Top-10 retrieval accuracy using each of the twelve layers of the original BERT to retrieve positive exemplars to solve the proxy task of EPR across ten tasks. Different colors represent different layers. Top-10 accuracy refers to the probability of retrieving the positive exemplar in the top 10 predictions. \textbf{Middle}: CKA scores between twelve layers of original BERT (x-axis) and the final layer of BERT of EPR trained on ten tasks. \textbf{Right}: CKA scores between each layer of the original BERT. These CKA scores are min-max normalized for better visualization. We use GPT-Neo \cite{gpt-neo} as the LLM.} 
    \vspace{-3ex}
    \label{fig:2}
\end{figure*}
Although semantic similarity generally excels in text retrieval, our observations show that low-level similarity (e.g., BM25) can sometimes outperform semantic counterpart in the demonstration selection task, especially on Nl2Bash~\cite{datasetnl2bash} and SWAG~\cite{dataset:swag}. Thus, we speculate that a critical aspect that makes the learning-based exemplar retriever effective lies in its ability to potentially learn and automatically integrate task-agnostic similarities of different levels during training (\HOne). In the following quantitative validation, we empirically find that the learning-based method EPR  dynamically ensembles the similarity encoded by various layers of an off-the-shelf BERT encoder.

As the first step, we validate the assumption that different layers of BERT, representing various levels of similarities, can exhibit different behaviors for different tasks as a retriever. To do that, following EPR, which builds a positive set $\{(x_i, x_i^+)\}_i^N$ using an LLM (as described in Sec.~\ref{sec:2.1}), we treat each $x_i^+$ as a gold exemplar to be retrieved for $x_i$. Then we utilize different layers of the original BERT (not fine-tuned) to retrieve/rank exemplars (as discussed in Sec.~\ref{sec:2.2}) for each $x_i$ and evaluate the Top-10 retrieval accuracy, representing the probability of retrieving the positive exemplar $x_i^+$ in top 10 predictions. The results on four tasks are depicted in Fig. \ref{fig:2}  when using GPT-Neo as the LLM. It reveals that different tasks exhibit distinct preferences towards specific layers, emphasizing different similarity levels. More information on these tasks can be found in Appendix~\ref{app:datasets}. Furthermore, while it is prevalent to employ the final layer of BERT for exemplar retrieval \cite{ACL22retriever1, ideal}, it is not consistently optimal, likely due to the potential inclusion of irrelevant information caused by BERT's pre-training tasks.

In the next step, we investigate what is encapsulated in the retriever learned by EPR. As this retriever utilizes the last layer of BERT to compute similarities, we extract the representations from this layer and compare those with representations from each layer of the original BERT to study the correlations between the EPR retriever and each original BERT layer. For this purpose, We introduce CKA \cite{CKA}, which effectively identifies correspondences between representations in different networks. Let $X^a \in \mathbb{R}^{n\times p_1}$ denote a matrix of activations of $p_1$ neurons for $n$ examples and $X^b \in \mathbb{R}^{n\times p_2}$ denote a matrix of activations of $p_2$ neurons for the same $n$ examples. The core insight of CKA lies in measuring the similarity between two matrices $X^a$ and $X^b$ by considering the inter-sample similarities. Specifically, CKA computes $K^a$ and $K^b$ to derive the inter-example similarity structures for $X^a$ and $X^b$, where $K^a =k^a(X^a,X^a)$, $K^b =k^b(X^b,X^b)$, $k^a$ and $k^b$ represent two linear kernels (i.e., ~$k(X, X) = X X^T$). Then, the CKA metric can be formulated as follows:
\begin{equation}
\small
\label{eq1}
\mathrm{CKA}(K^a, K^b)= \frac{\mathrm{HSIC(K^a,K^b)}}{\sqrt{\mathrm{HSIC}(K^a, K^a)\mathrm{HSIC}(K^b, K^b)}}, 
\end{equation}
where {\small $\mathrm{HSIC}(K^a,K^b)=\frac{1}{(n-1)^2}\text{tr}(K^aHK^bH)$} and {\small $\mathrm{HSIC}$}~is the aberration of Hilbert-Schmidt Independence Criterion. $H$ is the centering matrix ${H_{n}=I_{n}-{\tfrac {1}{n}}J_{n}}$ where $I_n$ is the identity matrix of size $n$ and $J_n$ is a $n$-by-$n$ all-ones matrix. 

 To measure the similarity between the last layer of the EPR retriever and each layer of the original BERT, we randomly sample $n=2000$ instances from the demonstration set $\mathcal{D}$ (If $\vert\mathcal{D}\vert <2000$, $n=\vert\mathcal{D}\vert$) for each task. Denote $X^{\textrm{EPR}}$ as the matrix composing $n$ rows of last-layer representations from the EPR retriever, and $X^l$ as the matrix composing $n$ rows of the $l$th-layer representations from the original BERT. Then we can calculate the CKA score $\mathrm{CKA}(K^{\textrm{EPR}}, K^l)$ for each task using Eq. \ref{eq1}. 

The results are depicted in Fig. \ref{fig:2} (Middle). Each row reflects the CKA similarity between the EPR retriever and each layer of the original BERT for a specific task. The CKA distribution across various tasks exhibits significant diversity among different BERT layers. This finding supports \HOne~that learning-based methods can adaptively aggregate multi-level (layer) similarities catering to different tasks. For instance, the results suggest that the exemplar retriever trained on Nl2Bash and SWAG tasks may prioritize low-level similarities, corroborating our experimental results where the BM25-based method outperforms higher-level semantic-based ones on these two datasets. Moreover, a similar validation using Llama 3~\cite{llama3} as the LLM is depicted in Fig.~\ref{fig:2:llma3} to evince \HOne~can generalize to more advanced LLMs.

\subsection{Output Similarity (\HTwo)}
\label{sec:3.2}
\begin{figure*}[tb]
    \centering
    \vspace{-3ex}
        \centering
        \includegraphics[width=\textwidth]{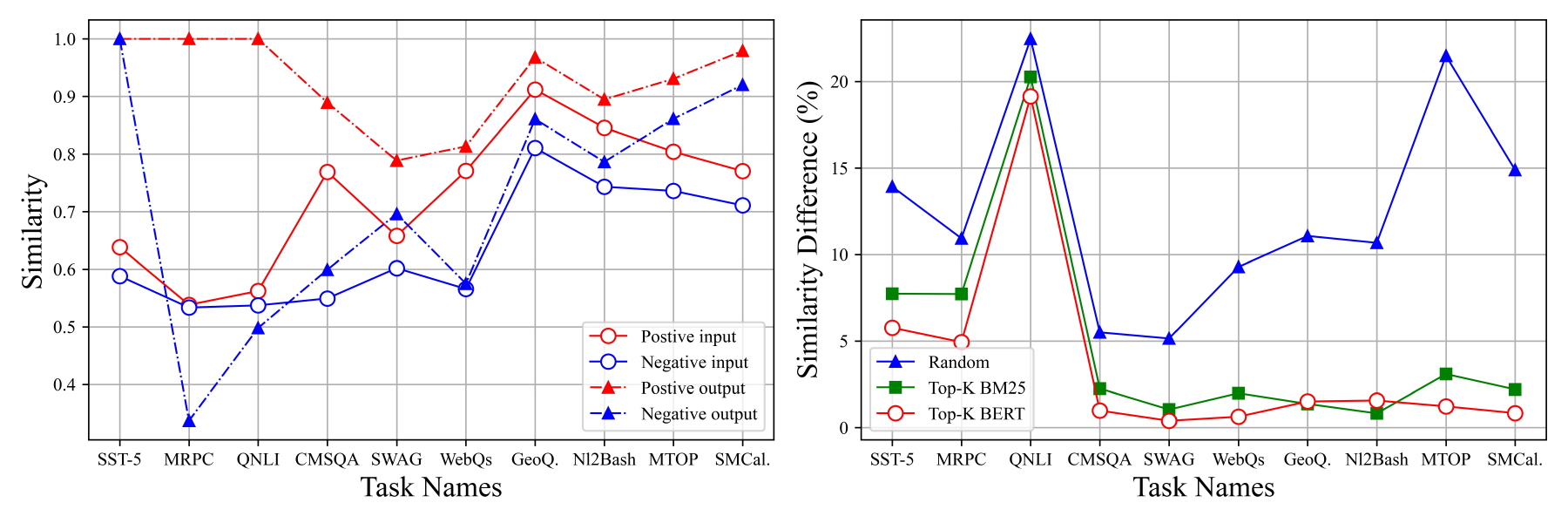}
        \vspace{-4ex}
    \caption{\textbf{Left}: Comparison of similarity between the input/output of positive and negative demonstration examples and the input/output of the test case across ten tasks for EPR. \textbf{Right}:  Difference between EPR and three task-agnostic demonstration exemplar selection methods in average similarity between the output of test case and retrieved exemplars. We use GPT-Neo \cite{gpt-neo} as the LLM.}
    \vspace{-3ex}
    \label{fig:3}
\end{figure*}
When employing the learning-based paradigm to acquire better similarity measurements between exemplars and test cases for ICL, such mechanics are expected to perform well on unseen tasks, given the high cost of data collection process. However, the sub-optimal generalization performance revealed by \citet{CEIL} suggests that the exemplar retriever, trained on the proxy task, primarily learns task-specific information. 

As data and training objectives can serve as a lens to analyze the behavior of neural network models, we first investigate the data generated for the proxy task involving positive and negative pairs, as shown in Fig. \ref{fig:3} (Left), which depicts the similarity between the input of positive/negative exemplars and the test case as well as the output of positive/negative exemplars and the test case. Specifically, for input similarity, we compute text similarity using  sentence-transformers\footnote{\url{https://huggingface.co/sentence-transformers/paraphrase-MiniLM-L6-v2}} for all tasks, while we compute exact match for the first three classification tasks and text similarity for other tasks as output similarity. Let $(x,y)$, $(x^{+}, y^{+})$, $(x^{-}, y^{-})$ denote the test case and corresponding positive and negative exemplars. The results indicate that the similarity between $x^{+}$ and $x$ is significantly higher than that between $x$ and $x^{-}$, affirming the efficacy of input similarity-based exemplar selection methods. Moreover, it is noteworthy that the similarity between $y^{+}$ and $y$ is also markedly higher than that between $y$ and $y^{-}$. 

Acknowledge that the training objective of the proxy task is to push the embeddings of $x$ and $x^{+}$ closer and push $x$ and $x^{-}$ away through contrastive learning in the embedding space. As a result, during the training phase, demonstration exemplars with similar outputs will resemble each other in this space due to the strong correlation between $y$ and $y^{+}$, which leads to a higher probability of selecting exemplars with outputs similar to the test case as prompts when the test case's output is unknown. Therefore,  we suggest that the success of learning-based approaches partly stems from the implicit prediction of the output of test cases during exemplar retrieval (\HTwo), which could be viewed as computing similarity of the joint distribution of input and output between the test case and exemplars.

After training on the proxy task, we utilize the EPR retriever to assess the similarity between the input of the test case and exemplars from the demonstration set to select top-K exemplars as prompts. To validate \HTwo, we evaluate the retriever's ability to learn the output similarity by computing the average similarity between the output of test cases and the retrieved exemplars. We compare EPR trained using GPT-Neo against task-agostic methods (i.e., Random, Top-K BM25 and Top-K BERT), as depicted in Fig. \ref{fig:3} (Right). We compute the output similarity for all tasks in the same way as experiments in  Fig. \ref{fig:3} (Left). The results show that the exemplar chosen by EPR has outputs more akin to the test case than other competitors, particularly in classification tasks where the output similarity can be well-captured by exact match. Similar validation is conducted for EPR with GPT2-XL and Llama 3 in Fig. \ref{fig:3:xl} and  Fig. \ref{fig:3:llma3}, and these results align with the above findings obtained for GPT-Neo, thus providing consistent support for \HTwo.

%% file: 4.method.tex
\section{Methodology}
\label{sec:method}
Building on the above findings, we propose two simple yet effective alternatives for learning-based demonstration exemplar section methods, which do not require costly interaction with LLMs to construct the proxy task. Specially, we introduce 1) Multi-level Similarity Maximization (\textbf{MLSM}) leveraging an adaptive ensemble of task-agnostic layer-wise anchors from BERT to achieve better cross-task generalization given \HOne; 2) Test Task Fine-tuning (\textbf{TTF}) infusing task-specific information to the retriever to enhance performances for this specific task according to \HTwo. 

\subsection{Multi-level Similarity Maximization (MLSM)}
\HOne~emphasizes that learning-based methods can adaptively integrate diverse similarities, which can be captured through different layers of a pretrained text encoder (e.g., BERT) from bottom to top. Inspired by ensemble learning \cite{ensemble1, ensemble2, ensembletta}, each layer can work as an expert for exemplar selection. The goal of \textbf{MLSM} is to integrate the insights from all experts by maximizing their agreement during the inference of LLMs.

However, as depicted in Fig. \ref{fig:2} (Right), each layer of BERT shows a high similarity to adjacent layers due to the residual design of transformers. Hence, we initially filter out redundant layers to avoid overfitting to the similarity of specific levels and reduce computational overhead. Specifically, given a task and its corresponding demonstration set $\mathcal{D}$, we sample a subset of unlabeled exemplars from $\mathcal{D}$ and compute layer-wise CKA scores between every pair of BERT layers, forming a similarity matrix $\mathbf{S}\in \mathbb{R}^{12\times12}$ where $S_{i,j}$ signifies the similarity between the $i$-th and $j$-th layers of BERT. We then employ an unsupervised K-means clustering algorithm to derive $n_l$ clusters, maximizing the intra-cluster CKA score while minimizing the inter-cluster CKA score, and designate the central node in each cluster as the representative layer. Finally, we attain a set of refined layers, denoted as $L = \{l_i\}^{n_l}_{i=1}$, as experts to represent the similarity at varying levels.

For a given test case $x^t$, we first sample a mini training set $\mathcal{D}_p=\{x_j\}^{n_p}_{j=1}$ and validation set $\mathcal{D}_v=\{x_j\}^{n_v}_{j=1}$ from $\mathcal{D}$. Then, for each layer $l_i \in L$, we compute the average of all token embeddings extracted from $l_i$ as the representation of $x^t$ (denoted as $\mathbf{h}^t$) and demonstration exemplars $x_j\in\mathcal{D}_p$ (denoted as $\mathbf{h}_j$). Following this, we compute the cosine similarity between $x^t$ and each exemplar in $\mathcal{D}_p$ as $\mathbf{r}_i=[\cos(\mathbf{h}^t, \mathbf{h}_1),...,\cos(\mathbf{h}^t, \mathbf{h}_{n_p})]$, and normalize it to obtain the probability distribution of these exemplars via $ \mathbf{e}_i = \mathrm{softxmax}(\frac{\mathbf{r}_i}{\tau})$ for layer $l_i$, where $\tau$ is the temperature parameter. Intuitively, such distribution can represent the ranking distribution of the demonstration exemplars when using the similarity level captured at $l_i$ for retrieval. After collecting the output distribution of all experts in $L$,  we aggregate them with learnable aggregation weights, denoted as $\mathbf{w}\in\mathbb{R}^{n_l}$ and get the ensembled exemplar ranking distribution as  $\hat{\mathbf{e}}=\mathrm{softmax}(\frac{\sum_{i=1}^{n_l}{w_i\mathbf{r}_i}}{\tau})$, where $\mathbf{w}$ is normalized before aggregation, i.e., $\sum_i^{n_l}w_i =1$. To encourage agreement among experts, we minimize the loss $\mathcal{L} = - \sum_{i=1}^{n_l}\hat{\mathbf{e}} \cdot \mathbf{e}_i$. The optimal $\mathbf{w}$ can be determined based on the loss on the validation set $\mathcal{D}_v$ by an early stopping strategy. Notably, this process does not rely on any task label (unsupervised) and focuses on general information regardless of specific tasks, thus catering to task-agnostic demands.

While \textbf{MLSM} focuses on the online scenario, where only one test point is observed during inference to align with the real-world demand~\cite{vs2023towards, ensembletta}, it can enable batch inference by updating $\mathbf{w}$ using a batch of test cases, thereby enhancing computational efficiency. 

\subsection{Test Task Fine-tuning (TTF)}
\HTwo~posits that the learning-based demonstration retriever inherently acquires the output similarity between exemplars and test cases for one specific target task when training on the proxy task. However, this proxy task requires costly interactions with LLMs for each target task to collect labeled positive/negative exemplars. To alleviate this issue, we propose \textbf{TTF} to infuse the output information to the retriever by fine-tuning it with additional modules customized for distinct tasks using labeled data from the demonstration set $\mathcal{D}$ directly. 

For convenience, let $f_{\bm{\theta}}$ denote the retriever and $q_{\bm{\phi}}$ denote the extra module, containing $\bm{\theta}$ and $\bm{\phi}$ as learnable paramters. For classification tasks,  $q_{\bm{\phi}}$ will be instantiated through various classification heads.  Given a test input $x$, assuming a linear classifier, the predication is derived by taking the argmax over the approximated probability distribution: 
\begin{equation}
\small
  \argmax\limits_{y_i}q_{\bm{\phi}}(Y=y_i|\mathbf{z})=\frac{\exp(\mathbf{z}\cdot\bm{\phi}_i)}{\sum_j \exp(\mathbf{z\cdot\bm{\phi}_j})}, 
\end{equation}
where $\mathbf{z}=f_{\bm{\theta}}(x)$ and $\bm{\phi}_i$ is the $i$-th component of the weights $\bm{\phi}$ corresponding to label $y_i$. As the prediction is determined by evaluating the distance between $\bm{\phi}_i$ and $\mathbf{z}$, test cases with a similar output are more likely to exhibit a smaller distance in the semantic space, as they are closer to their corresponding $\bm{\phi}$. Furthermore, previous research \cite{clssifer1, clssifer2} has leveraged $\mathbf{z}$ as a pseudo-prototype for each label to construct non-parametric classifiers, providing evidence that \textbf{TTF} can effectively encapsulate the input-output relationship for classification.

For generation tasks, while decoder-only frameworks are unsuitable for deriving sentence embeddings without prompting or fine-tuning \cite{sgpt}, we adopt the encoder-decoder architecture, where $q_{\bm{\phi}}$ is instantiated by the decoder and the retriever $f_{\bm{\theta}}$ works as the encoder. Since the decoder generates new tokens based on the encoder's output, allowing the encoder's output to capture pertinent input-output information naturally, we follow \citet{DBLP:conf/acl/NiACMHCY22} to use the average pooling of all token embeddings extracted from the last layer of the encoder to represent test cases and exemplars. The detailed implementation of \textbf{TTF} can be found in Appendix. \ref{appedix:impdetail}. Ultimately, \textbf{TTF} acquires the output similarity between demonstration exemplars and test cases by training the retriever on the demonstration set $\mathcal{D}$, thereby adapting to task-specific requirements. 

% \begin{algorithm}[t]
%    \caption{Demonstration selection}
%    \label{alg:demo-selection}
% \begin{algorithmic}
% \STATE {\bfseries Input:} dataset $\mathcal{D}^d$ for a task $d$. LLM with fine-tuned concept tokens $M'$. The number of demonstrations $k$.
% \STATE {\bfseries Output:} A set of selected demonstrations.
% \FOR{each $(X^d, Y^d)$ in $\mathcal{D}^d$}
%     \STATE Compute $\hat{P}_M^d(\hat{\theta}^d|X^d, Y^d)$;
% \ENDFOR
% \STATE Select top $k$ examples with the largest $\hat{P}_M^d(\hat{\theta}^d|X^d, Y^d)$, denoted as $(X^d_1, Y^d_1), ..., (X^d_k, Y^d_k)$;
% \end{algorithmic}
% \end{algorithm}

%% file: 5.exp.tex
\section{Experiments}
\label{sec:5}
\input{tabs/datasets}
\paragraph{Datasets.} We conduct experiments on ten datasets spanning seven categories of NLP tasks: sentiment analysis, paraphrase detection, natural language inference, commonsense reasoning, open-domain question answering, code generation and semantic parsing. As certain datasets lack a test set, we take the training split as the demonstration set and the validation split for evaluation across all datasets. The statistics of all datasets are listed in Table \ref{tab:datasets}. A detailed description of these datasets and prompts to reproduce our experimental results are shown in Appendix \ref{app:datasets}.

\paragraph{Baselines.} In line with previous studies \cite{epr, CEIL, li-etal-2023-unified}, we consider two baseline categories based on whether to use labeled data in the demonstration set: unsupervised and supervised methods. The unsupervised category includes \textsc{Random}, which randomly selects exemplars from the demonstration set without repetition; \textsc{Top-K BM25}, which employs BM25 \cite{bm25} to retrieve the Top-K most similar exemplars based on low-level text similarity; and \textsc{Top-K BERT}, which generates text representations by averaging token embeddings from the final layer of BERT \cite{bert} and retrieves the Top-K most similar exemplars based on semantic similarity. The supervised category includes \textsc{EPR} \cite{epr}, which utilizes Top-K BM25 to generate demonstration candidates and scores them using LLMs to construct a proxy task, subsequently fine-tuning BERT in \textsc{Top-K BERT} using this task; and \textsc{CEIL} \cite{CEIL}, which employs EPR to generate demonstration sequence candidates, scores them using LLMs to construct a proxy task, and further fine-tunes BERT using this task. \textsc{CEIL} balances diversity and relevance using a trade-off parameter and searches for the optimal exemplar combination using Determinantal Point Processes \cite{dpp}. While mainly utilizing BERT as the retriever of \textbf{MLSM} and \textbf{TTF}, we exploit T5 for \textbf{TTF} on the generation tasks because BERT-based encoder-decoder models cannot handle generation tasks\footnote{That is because of the random initialization of external cross attention modules and the lack of sufficient training data for BERT-based generation models. }. 
The implementation detail of our methods and all baselines can be found in Appendix~\ref{appedix:impdetail}.
\begin{table}[tb]
\centering
\caption{Main results on the classification task. $\clubsuit$ indicates methods requiring costly interaction with LLMs.} %, and $\spadesuit$ denotes no such requirement.}
\label{man:cls}
\begin{adjustbox}{width=\linewidth, center}
\begin{tabular}{lcccccc}
\toprule
Method & SST-5 & MRPC& QNLI & CMSQA & SWAG & Avg. \\
\midrule
\textit{Unsupervised} &  &  &  &  &  &  \\
Random & 28.61	 & 65.93 & 55.08	 &42.34 & 41.39	 & 46.67 \\
Top-K BM25 & 32.06 & 65.93 & 60.11	 &35.79	 & 43.35 & 47.45 \\
Top-K BERT & 32.70	 & 69.12 & 60.94 &35.87	 & 41.09	& 47.94\\ 
\cdashline{2-7}[2pt/3pt]
MLSM & 33.15 & 69.87 & 65.02	 &37.26	 & 41.49 & \textbf{49.36}\\ 
\midrule
\textit{Supervised}  &  &  &  &  &  &  \\
EPR$^\clubsuit$ & 36.88	 & 81.37 & 77.87	 &38.74 & 43.39 & 55.65 \\
CEIL$^\clubsuit$ & 37.69	 & 77.94	 & 80.58	 &38.90 & 43.84	& 55.79\\
\cdashline{2-7}[2pt/3pt]
TTF & 42.14	 & 74.51	 & 85.08	 &47.83	 & 55.72	& \textbf{61.06}\\ 
% MLM  & 32.70	 & 69.12 & 60.94 &35.87	 & 41.09	& 47.94\\ 
% Whitening  & 32.70	 & 69.12 & 60.94 &35.87	 & 41.09	& 47.94\\ 
\bottomrule
\end{tabular}
\end{adjustbox}
\vspace{-3ex}
\end{table}
\paragraph{Experiment settings.} Following \citet{CEIL}, we employ GPT-Neo (2.7B) \cite{gpt-neo} as the main LLM in this study and conduct experiments on a smaller GPT-2 XL \cite{gpt2xl} (1.5B) and text-davinci-002 to verify the transferability of our methods. Furthermore, we extend our experiments to more advanced LLMs, including Llama 3 (8B) and GPT-3.5, to support our hypotheses and findings in Appendix \ref{app:advancedllm}. Due to computational constraints and different maximum context sizes among LMs, we restrict the number of in-context exemplars to 20. These exemplars are sorted based on their similarities to test cases in ascending order following prior practices \cite{epr, emnlp23skillselection1, ACL22retriever1}. For model evaluation, we compare the predicted output with ground truth for all methods and report Accuracy (Acc.) and Exact Match (EM) for classification and generation tasks, respectively. 
\begin{table}[tb]
\centering
\caption{Main results on the generation tasks. $\clubsuit$ indicates methods requiring costly interaction with LLMs.}
\label{man:gen}
\begin{adjustbox}{width=\linewidth, center}
\begin{tabular}{lcccccc}
\toprule
Method & WebQs & GeoQ.& NL2B. & MTOP & SMCal. & Avg. \\
\midrule
\textit{Unsupervised} &  &  &  &  &  &  \\
Random & 3.79	 & 25.36 & 31.27 &3.98 & 3.70	 & 13.62 \\
Top-K BM25 & 14.17 & 65.71 & 58.81 &49.66	 & 44.02 & 46.48 \\
Top-K BERT & 14.17	& 64.64 & 52.45 &51.36 & 44.76	& 45.48\\ 
Top-K T5 & 16.24 & 70.35 & 43.29 & 53.02	& 42.83& 45.14\\ 
\cdashline{2-7}[2pt/3pt]
MLSM  &16.14 & 68.93 & 56.11	 &54.05 & 47.72 & \textbf{48.59}\\ 
\midrule
\textit{Supervised}  &  &  &  &  &  &  \\
EPR$^\clubsuit$ & 17.62		 & 73.21	& 77.87	 &60.82	& 60.49	 & \textbf{53.43}\\
CEIL$^\clubsuit$ & 17.08 & 70.71	& 53.66	&63.40	& 56.30& 52.23\\
\cdashline{2-7}[2pt/3pt]
% TTF  & 10.97 & 47.86	& 28.69	&31.59 & 13.36& 26.49\\ 
TTF (T5) & 17.07 & 71.43 & 46.30	 &58.12	 & 51.06 & 48.80\\ 
\bottomrule
\end{tabular}
\end{adjustbox}
\vspace{-3ex}
\end{table}
\paragraph{Main Results.} We compare \textbf{MLSM} and \textbf{TTF} with existing unsupervised (using off-the-shelf models directly) and supervised learning-based baselines on classification tasks (Table \ref{man:cls}) and generation tasks (Table \ref{man:gen}). The results show that \textbf{MLSM} consistently outperforms all unsupervised baselines in most cases, achieving an average improvement of 1.42\% over the best baseline, Top-K BERT (semantic similarity), on classification tasks, and an average improvement of 2.11\% over the best baseline, Top-K BM25 (low-level similarity), on generation tasks. This suggests that while different similarities excel at different tasks, \textbf{MLSM} can adaptively integrate multi-level similarities for various tasks by updating the aggregation weight of the experts for each test case, thus providing evidence for \HOne. Moreover, supervised methods generally show a clear advantage over \textbf{MLSM} across all tasks, highlighting the benefit of incorporating task-specific information into the retriever. Notably, despite avoiding costly integration with LLMs, \textbf{TTF} surpasses both EPR and CEIL, achieving over 5\% absolute improvements on classification tasks, and consistently outperforms \textbf{MLSM} across all generation tasks except NL2Bash. It suggests that test task fine-tuning can be a more effective alternative to constructing proxy tasks in resource-limited scenarios,  further validating \HTwo. However, \textbf{TTF} underperforms compared to EPR and CEIL on some generation tasks likely due to inherent limitations of the encoder-decoder framework in retrieval tasks, particularly in identifying which parts of the model capture relevant input-output information. For instance, Top-K T5 performs worse than Top-K BERT regarding average accuracy across all generation tasks. 

\begin{figure}[tb]
    \centering
    \vspace{-1ex}
        \centering
\includegraphics[width=\linewidth]{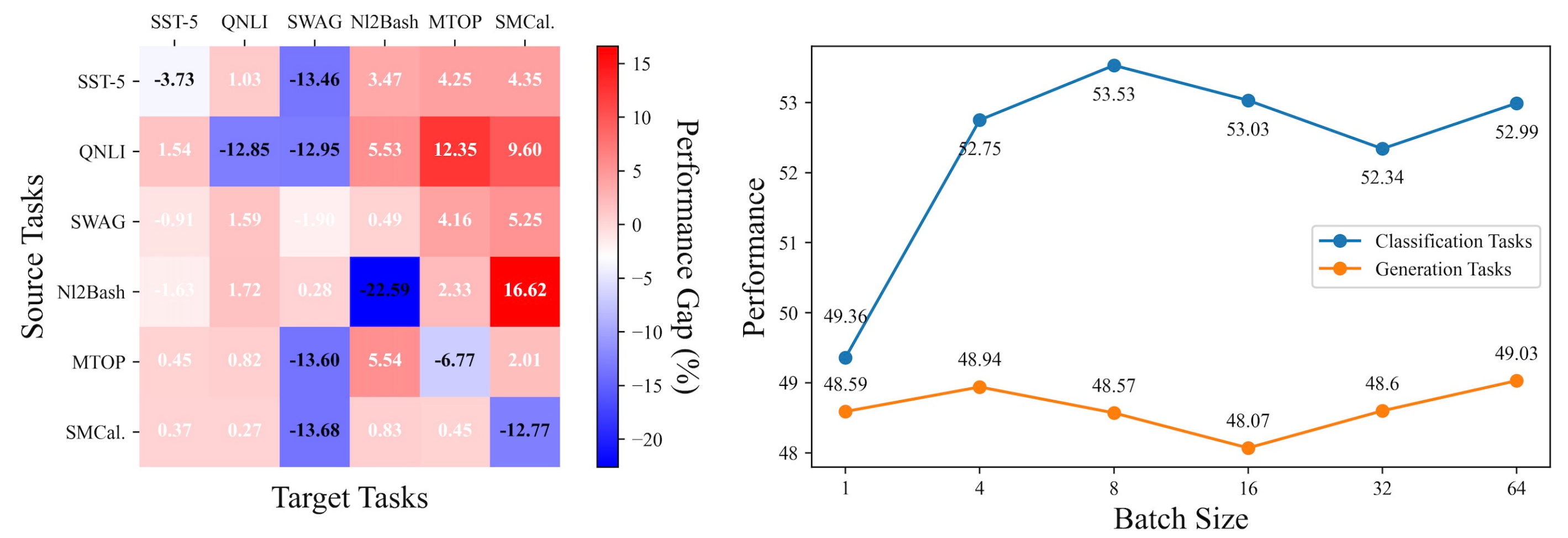}
        % \vspace{-6ex}
    \caption{\textbf{Left}: Comparison of transferability between EPR and MLSM. We show the absolute improvement of MLSM over EPR. \textbf{Right}: Comparisons of different batch sizes for MLSM. } 
    \vspace{-2ex}
    \label{fig:4}
\end{figure}

\paragraph{Transfer across Tasks for MLSM.} We compare EPR and \textbf{MLSM} on cross-task experiments, where EPR is trained on a source task and transferred to a target task, as depicted in Fig. \ref{fig:4} (Left). The results show that EPR generally performs worse than \textbf{MLSM}, particularly when transferring between classification and generation tasks. It suggests that learning-based exemplar selection methods overfit task-specific features when trained on the proxy task, making it challenging to justify the high cost of data collection. In contrast, \textbf{MLSM} is a practical solution for task-agnostic demands, as it only leverages information from the test case to adapt to different tasks during LLM inference.

\paragraph{Ablation of Batch Size for MLSM.}  While \textbf{MLSM} assumes only a single test case is available for learning the aggregation weight $\mathbf{w}$ of different similarity levels, we perform an ablation study to assess the impact of increasing batch size in Fig. \ref{fig:4} (Right). The results indicate that \textbf{MLSM} generally benefits from a larger batch size, especially on classification tasks, showing over 4\% average improvements when the batch size is 8.  This improvement can be attributed to the fact that test cases in the same batch tend to share common patterns of multi-level similarities (i.e., similar $\mathbf{w}$), further suggesting that multi-level similarities are essential for selecting good demonstration exemplars.

\paragraph{Transfer across LLMs.} 
\begin{table}[]
\centering
\caption{Results of cross-LLM transferability validation. We show the absolute \textbf{improvement} of TTF and MLSM over Top-K BERT. }
\label{man:crossllm}
\begin{adjustbox}{width=\linewidth, center}
\begin{tabular}{lccccc|ccccc}
\toprule
  & \multicolumn{5}{c}{\textbf{TTF}}                                         & \multicolumn{5}{c}{\textbf{MLSM}}                                            \\
\midrule
LLM        & SST-5 & MRPC   & QNLI & CMSQA  & Avg. & SST-5 & MRPC    & GeoQ.  & NL2Bash & Avg. \\

GPT-2 XL (1.5B)  & 3.54         & 0.00 & 5.35        & 6.38 & 3.82       & 1.54         & 0.00 & 1.07 & 4.57           & 1.74       \\
GPT NEO (2.7B) & 9.45         & 5.39 & 24.14        & 11.96 & \textbf{12.73}       & 0.05         & 0.75  & 4.29 & 3.66           & 2.29       \\
text-davinci-002  & 3.27         & 1.51 & 18.52        & 1.15 & 6.11       & 1.82         & 1.47  & 3.21 & 3.02           & \textbf{2.38}      \\
\bottomrule
\end{tabular}
\end{adjustbox}
\vspace{-2ex}
\end{table}
We validate the versatility of \textbf{TTF} and \textbf{MLSM} on GPT-2 XL, GPT-NEO, and text-davinci-002 in Table \ref{man:crossllm}. The results indicate that both methods can enhance ICL performance across different LLMs. \textbf{TTF} consistently outperforms \textbf{MLSM}, verifying the effectiveness of acquiring task-specific output similarity between exemplars and test cases. However, \textbf{TTF} exhibits higher variance in performance across different LLMs than \textbf{MLSM}, suggesting different LLMs have varying abilities to exploit exemplars with similar outputs to the test case. Additionally, the better performance of \textbf{TTF} on GPT-NEO compared to text-davinci-002 implies that the latter's stronger ability may make it more resilient to prompt choices.

%% file: tabs/datasets.tex
% \usepackage{multirow}
\begin{table*}[t]
\centering
\caption{The statistics of ten datasets. We report the number of training instances after deduplicating.}
\label{tab:datasets}
% \vskip 0.15in
\scalebox{0.85}{
\begin{tabular}{lllcc}
\toprule
\textbf{Type} & \textbf{Dataset} & \textbf{Task} & \textbf{Train} & \textbf{Validation}\\
\midrule
\multirow{5}{*}{{Classification}} 
 & SST-5~\citep{datasetsst5} & Sentiment Analysis & 8,534 & 1,101\\
 & MRPC~\citep{dataset:mrpc} & Paraphrase Detection & 3,668 & 408\\
 & QNLI~\citep{dataset:qnli} & Natural Language Inference & 104,707 & 5,463\\
 & CMSQA~\citep{dataset:cmsqa} & Commonsense Reasoning & 9,740 & 1,221\\
 & HellaSwag~\citep{dataset:swag} & Commonsense Reasoning & 52,611 & 20,006\\
 \midrule
\multirow{5}{*}{{Generation}}
 & WebQs~\citep{dataset:webqs} & Open-Domain QA & 3,778 & 2,032\\
 & GeoQuery~\citep{dataset:gequery} & Code Generation & 404 & 280 \\
 & Nl2Bash~\citep{datasetnl2bash} & Code Generation & 7,441 & 609 \\
 & MTOP~\citep{dataset:mtop} & Semantic Parsing & 15,564 & 2,235 \\
 & SMCalFlow~\citep{smcalflow} & Semantic Parsing & 102,491 & 14,751 \\
 \bottomrule
\end{tabular}}
\vspace{-3ex}
\end{table*}

%% file: 7.conclusion.tex
\section{Conclusion}
\label{sec:conclusion}
In this work, we delve into the mechanism of learning-based demonstration exemplar selection methods. We speculate the advantages of these methods stem from their ability to integrate similarities of different levels for exemplar selection (\HOne) and their capacity to choose exemplars with similar outputs to the test case (\HTwo). Motivated by these hypotheses, we introduce two simple but effective exemplar selection methods, \textbf{MLSM} and \textbf{TTF}, tailored to task-agnostic and task-specific demands without costly interactions with LLMs. Quantitative validations and the effectiveness of both methods provide substantial evidence for \HOne~and \HTwo. In summary, our work offers insights into more efficient LLM deployment in practical applications and may benefit transparent research on exemplar selection methods and ICL. 
\newpage
\section*{Limitations.}
In this section, we discuss two technical limitations of our work.

\noindent\textbf{Combination of MLSM and TTF:} Based on our two findings related to the working mechanism of learning-based exemplar section methods, we propose two cost-effective selection approaches: \textbf{MSLM} maximizing the agreement across the similarities of different levels and \textbf{TTF} fine-tuning a retriever with labeled data from the demonstration set to learn task-specific similarity between the output of exemplars and test cases. While \textbf{MSLM} and \textbf{TTF} excel in task-agnostic and task-specific scenarios, combining them could potentially further enhance task-specific performance. To investigate this, we replace the original BERT in \textbf{MSLM} using the trained retriever in TTF and conduct experiments on five classification tasks using the same implementation detailed in Appendix \ref{appedix:impdetail}. As shown in Table \ref{tab:combination}, although the combination of both methods significantly outperforms \textbf{MSLM} with an average improvement of around 4\%, it falls short of \textbf{TTF} by over 6\%. This performance drop suggests that the similarity between the output of exemplars and test cases is superior to similarities from other layers, and while TTF's final layer effectively captures such task-specific output similarity, integrating it with other sub-optimal ones could introduce noise, negatively impacting the exemplar selection for ICL.

\noindent\textbf{Better Implementation of \HTwo~than TTF:} In \HTwo, we empirically find that the success of learning-based methods partially stems from their ability to choose the demonstration exemplar with similar output to the test case. We propose \textbf{TTF} to simulate such output-based similarity by implicitly learning task-specific information from labeled demonstration exemplars using different task heads. Despite showing promise on classification tasks, \textbf{TTF} is ineffective for generation tasks compared to EPR and CEIL. We attribute this to 1) the difficulty in identifying model components that capture effective input-output relationships in a decoder-encoder framework and 2) the need for extensive data to fine-tune generation task heads or more advanced pre-trained models.

To further explore \HTwo, we try two approaches: First, akin to EPR, we select the exemplar with the most similar output to the test case as a positive pair and the most dissimilar one as a negative pair and then fine-tune a retriever. However, this led to a performance collapse, likely due to the complexity of modeling nuanced input-output similarities for generation tasks. Secondly, building upon \textbf{TTF}, we generate outputs using T5 for each test case and compute similarities between inputs and outputs of demonstration exemplars and test cases. Finally, we integrate these similarities with a predefined ratio (0.9 and 0.1), yielding an average improvement of 1\% over \textbf{TTF}. However, this method requires first generating answers for test cases, making it less efficient than using input embedding for exemplar retrieval. Additionally, recent exemplar selection methods \cite{emnlp23skillselection1, zhou2024llms, sunemnlp2024retrieved} that use LLMs to briefly describe the reasoning process and compute the similarity between such descriptions of exemplars and test case for retrieval, can be seen as an instantiation of \HTwo, as they also implicitly model the input-output relationship.

In summary, the main contribution of our work lies in suggesting and validating two hypotheses regarding learning-based exemplar selection methods. While \textbf{MSLM} and \textbf{TTF} show advantages over existing demonstration exemplar section methods, they are just two possible implementations of our findings. More advanced exemplar selection methods could be developed based on these insights. As a result, we advocate for further research in this area to enhance the efficient deployment and transparency of LLMs and ICL.
\input{tabs/tab.limitation}

\section*{Ethics Statement}
This paper adheres to the ACM Code of Ethics and Professional Conduct. This work presents two key findings about the working mechanism of learning-based demonstration selection and two methods for low-cost exemplar selection, which do not pose any societal harm. All datasets used are publicly available. We will release our code following the licenses of any utilized artifacts.

%% file: tabs/tab.limitation.tex
\begin{table}[bt]
\centering
\caption{Experimental results for the combination of MLSM and TTF}
\label{tab:combination}
\begin{adjustbox}{width=\linewidth, center}
\begin{tabular}{lcccccc}
\toprule
Method & SST-5 & MRPC& QNLI & CMSQA & SWAG & Avg. \\
\midrule
MLSM & 33.15 & 69.87 & 65.02	 &37.26	 & 41.49 & 49.36\\ 
TTF & 42.14	 & 74.51	 & 85.08	 &47.83	 & 55.72	& 61.06\\ 
TTF and MLSM  & 36.14 & 71.07 & 65.31	 &45.61	 & 50.27 & 53.69\\ 
\bottomrule
\vspace{-7ex}
\end{tabular}
\end{adjustbox}
\end{table}

%% file: 6.appendix.tex
\appendix
\section{Outline of the Appendix}
The appendix is organized as follows: Appendix \ref{app:com-exp} provides descriptions of the datasets used in our experiments, the prompts for reproducing our work, and the implementation details for all baselines as well as our proposed \textbf{MLSM} and \textbf{TTF} methods. Appendix \ref{app:advancedllm} examines the generalization of our findings and methods to more advanced LLMs. Appendix \ref{app:icl} offers qualitative validation of \HOne~and \HTwo~by connecting our results with existing explanatory work on ICL. Appendix \ref{app:Significance} presents the statistical significance of our proposed methods. Appendix \ref{app:foundation} outlines the theoretical foundation of \textbf{MLSM} and \textbf{TTF}, while Appendix \ref{app:weightanalysis} analyzes the aggregation weights in \textbf{MLSM}. Lastly, Appendix \ref{app:compcost} discusses the running efficiency of both methods.
\section{Experimental Setup}
\label{app:com-exp}
\begin{table*}[htb]
\centering
\caption{Datasets with corresponding prompts and examples used in the experiments.}
\label{tab:examples}
\includegraphics[width=\textwidth]{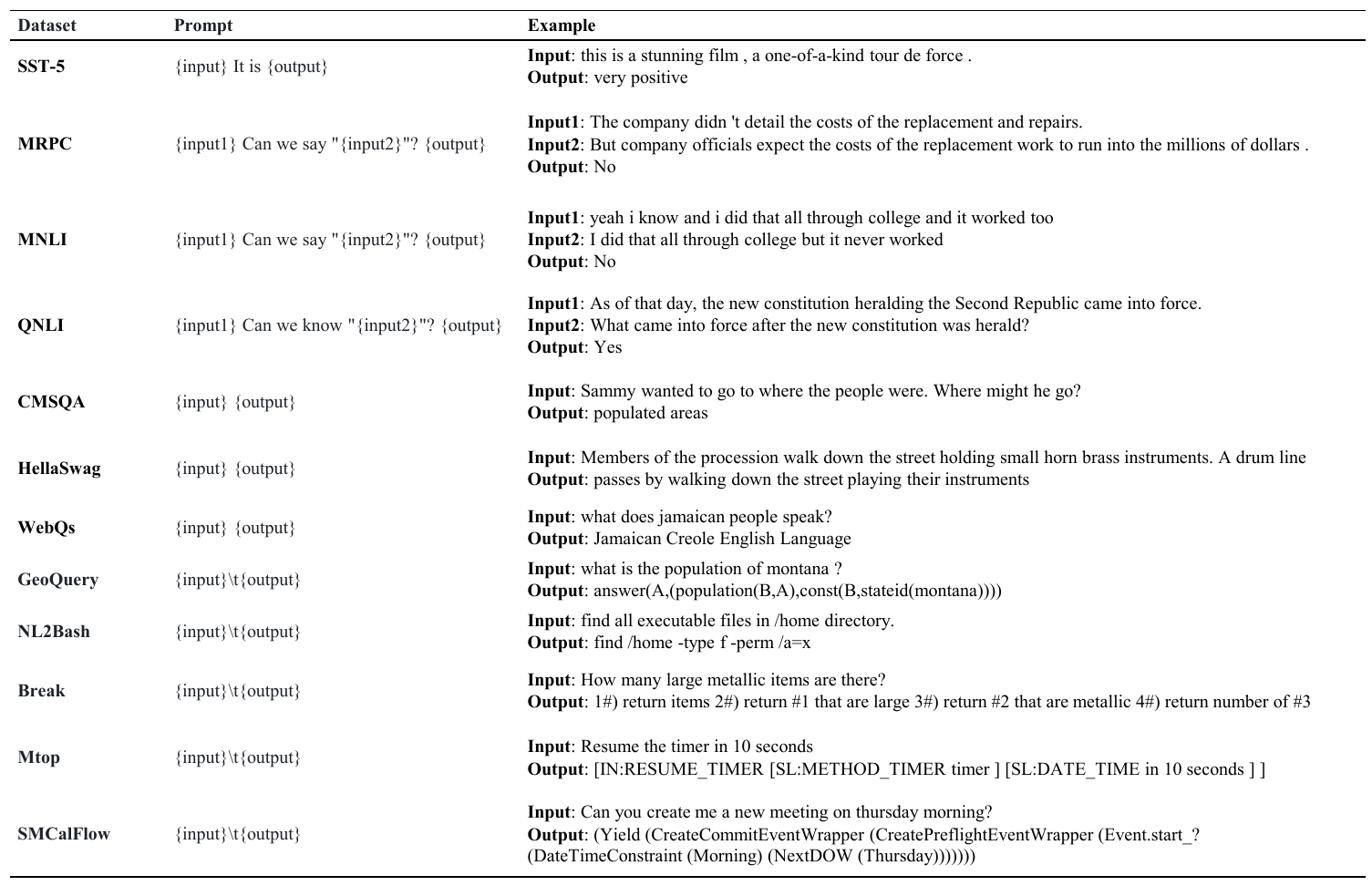}
\end{table*}

\subsection{Datasets} 
\label{app:datasets}
Following existing work \cite{CEIL}, we conduct experiments on five classification tasks and five generation tasks\footnote{We exclude MNLI \cite{datasetmnli} to reduce computation cost and Break \cite{break} because of failure to reproduce its evaluation method.}. While we advise readers to refer to the detail of each dataset in the original work \cite{CEIL}, we provide the prompts and examples for each dataset in Table~\ref{tab:examples} and offer a detailed description of each dataset below for completeness.

\paragraph{SST-5~\cite{datasetsst5}} is a sentiment classification benchmark containing five fine-grained classes including `very positive', `positive' `neutral', `negative', and `very negative'. 
\paragraph{MRPC~\cite{dataset:mrpc}} is a corpus of sentence pairs automatically extracted from online news sources, with human annotations for whether the sentences in the pair are semantically equivalent. 

\paragraph{MNLI~\cite{datasetmnli}} is a crowdsourced collection of sentence pairs with textual entailment annotations. Given a premise sentence and a hypothesis sentence, the task is to predict whether the premise entails the hypothesis (entailment), contradicts the hypothesis (contradiction), or neither (neutral). 

\paragraph{QNLI~\cite{dataset:qnli}} is a question-answering dataset consisting of question-paragraph pairs, and the task is to determine whether the context sentence contains the answer to the question.
\paragraph{CMSQA~\cite{dataset:cmsqa}} (short for CommonsenseQA) is a multiple-choice question-answering dataset that requires different types of commonsense knowledge. The task is to predict the correct answer out of five provided candidate answers. 

\paragraph{HellaSwag~\cite{dataset:swag}} is a large-scale dataset of grounded commonsense reasoning.
Each question has four candidate answers: a video caption from ActivityNet Captions~\cite{heilbron2015activitynet} and the Large Scale Movie Description Challenge~\cite{rohrbach2017movie}. The three incorrect answers are adversarially generated and human-validated to deceive machines. The correct answer is the actual video caption for the subsequent occurrence in the video.

\paragraph{WebQs~\cite{dataset:webqs}} is question-answer pairs obtained from the web. The questions are selected using Google Suggest API,
and the answers are entities in Freebase.

\paragraph{Nl2Bash~\cite{datasetnl2bash}} is a dataset for the problem of mapping English sentences to Bash commands. The corpus consists of text–command pairs, where each
pair consists of a Bash command scraped from the web and an expert-generated natural language description.

\paragraph{GeoQuery~\cite{dataset:gequery}} contains a parallel
corpus of 880 English questions about US geography paired with Prolog queries.

\paragraph{Break~\cite{break}} is a dataset that maps complex natural language questions into a language-based meaning representation. The question is decomposed into an ordered list of atomic steps used as the target sequence. We use the low-level Break subset following~\citep{epr}.

\paragraph{MTOP~\cite{dataset:mtop}} is a multilingual task-oriented semantic parsing
dataset covering six languages and 11 domains. The target commands are complex queries featuring nested intent-slot prediction.
Similar to past work ~\cite{epr}, we use the English subset of MTOP.

\paragraph{SMCalFlow~\cite{smcalflow}} is a large dialogue dataset featuring natural conversations about tasks involving calendars, weather, places, and people. The meaning representation is an executable dataflow program featuring API calls, function composition, and complex constraints. 

\subsection{Implementation Details}
\label{appedix:impdetail}
We employ the implementation\footnote{\url{https://github.com/HKUNLP/icl-ceil}} from \citet{CEIL} for all baselines. Specifically, for \textsc{EPR} and \textsc{CEIL}, we limit the maximum instances in the proxy task to 4,000 ($\vert\mathcal{D}^s\vert=4,000$) and sample 50 candidates for each instance to create positive and negative pairs. It is worth noting that collecting these data for both methods (i.e., 200,000 queries to LLMs) is pretty expensive and time-consuming, especially for CEIL, where each candidate sequence involves 16 exemplars. 

For our proposed \textbf{MLSM}, we randomly sample 1,000 examples ($n_c=1,000$) from the demonstration set $\mathcal{D}$ to compute layer-wise CKA scores and obtain three representative layers through clustering ($n_l=3$). Then, we randomly sample 256 and 64 examples ($n_t=256$ and $n_v=64$) for each test case from $\mathcal{D}$ as mini training and validation sets, respectively. The temperature of the softmax function is set to 0.01 ($\tau=0.01$). We utilize Adam optimizer with batch size 32 and learning rate 0.1 to learn the aggregation weight $\mathbf{w}$ in fewer epochs. 
 
For our proposed \textbf{TTF}, we utilize the labeled data from the demonstration set, consisting of input-output pairs ($(x,y)$), to train the retriever with customized heads tailored to different tasks, where the goal is to predict $y$ given $x$.  We instantiate $f_{\bm{\theta}}$ with BERT and $q_{\bm{\phi}}$ with different task heads for classification tasks. Concretely, for SST-5, MRPC and QNLI, we utilize the sequential classification head\footnote{\url{https://huggingface.co/docs/transformers/model_doc/bert\#transformers.BertForSequenceClassification}} and train the model using Adam optimizer with batchsize 32, learning rate 5e-4 and weight decay 1e-4. For SWAG and CMSQA, we adopt the multi-choice head\footnote{\url{https://huggingface.co/docs/transformers/model_doc/bert\#transformers.BertForMultipleChoice}} and also train the model using Adam optimizer with batchsize 8, learning rate 5e-4 and weight decay 1e-4. Additionally, for generation tasks, we instantiate $f_{\bm{\theta}}$ and $q_{\bm{\phi}}$ using the encoder and decoder of T5\footnote{\url{https://huggingface.co/google-t5/t5-base}} and utilize to Adam optimizer with batchsize 8, learning rate 4e-5 and weight decay 0.01. 

Furthermore, we conducted all experiments for \textsc{EPR} and \textsc{CEIL} on two NVIDIA A100 GPUs (40GB), while the remaining experiments were performed on two NVIDIA V100 GPUs (30GB). Each main experiment is repeated three times using different random seeds to mitigate the effects of randomness.

\section{Experiments on Advanced LLMs}
\label{app:advancedllm}
\begin{figure*}[tb]
    \centering
    \vspace{-1ex}
        \centering
        \includegraphics[width=\linewidth]{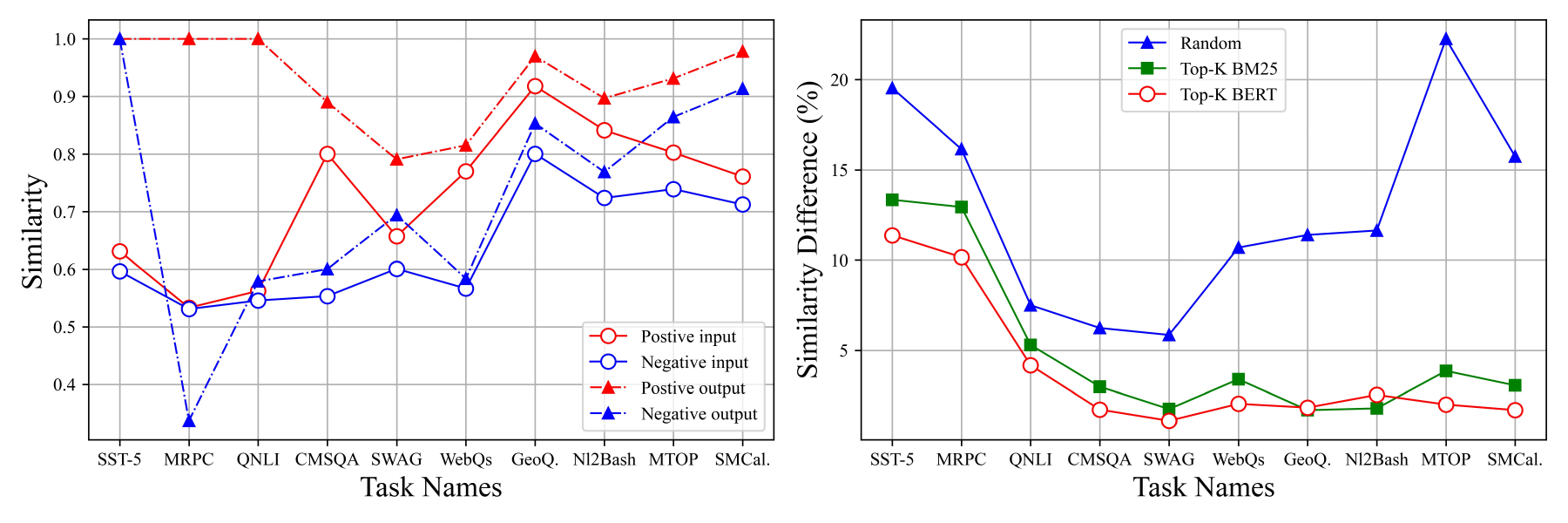}
        % \vspace{-6ex}
    \caption{\textbf{Left}: Comparison of similarity between the input/output of positive and negative demonstration examples and the input/output of the test case across ten tasks for EPR. \textbf{Right}: Difference between EPR and three task-agnostic demonstration exemplar selection methods in average similarity between the output of test case and retrieved exemplars. We use GPT-2 XL \cite{gpt-neo} as the LLM.}
    \label{fig:3:xl}
\end{figure*}

\begin{figure}[tb]
    \centering
    % \vspace{-1ex}
        \centering
        \includegraphics[width=\linewidth]{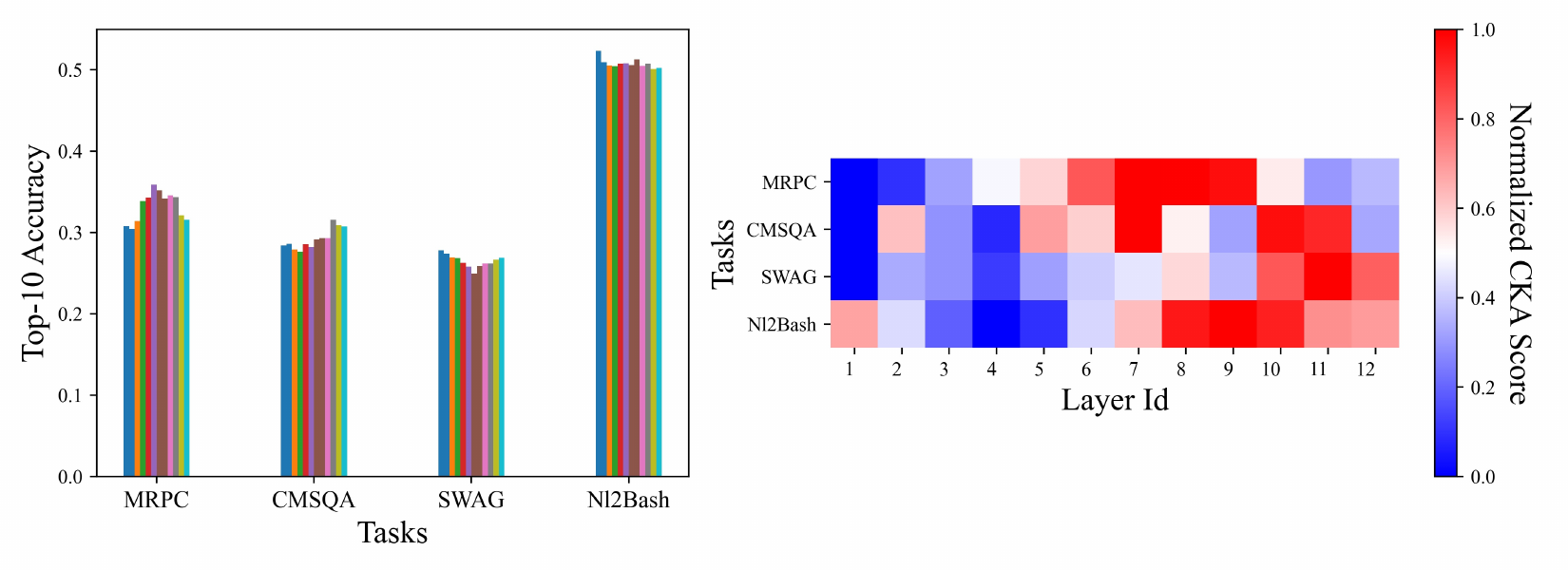}
        % \vspace{-6ex}
    \caption{ \textbf{Left}: Top-10 retrieval accuracy using each of the twelve layers of the original BERT to retrieve positive exemplars to solve the proxy task of EPR across four tasks. Different colors represents different layers. Top-10 accuracy refers to the probability of retrieving the positive exemplar in the top 10 predictions. \textbf{Middle}: CKA scores between twelve layers of original BERT (x-axis) and the final layer of BERT of EPR trained on four tasks. We use Llama3 (8B) as the main LLM.}
        \vspace{-3ex}
 \label{fig:2:llma3}
\end{figure}

\begin{figure}[tb]
    \centering
    % \vspace{-1ex}
        \centering
        \includegraphics[width=\linewidth]{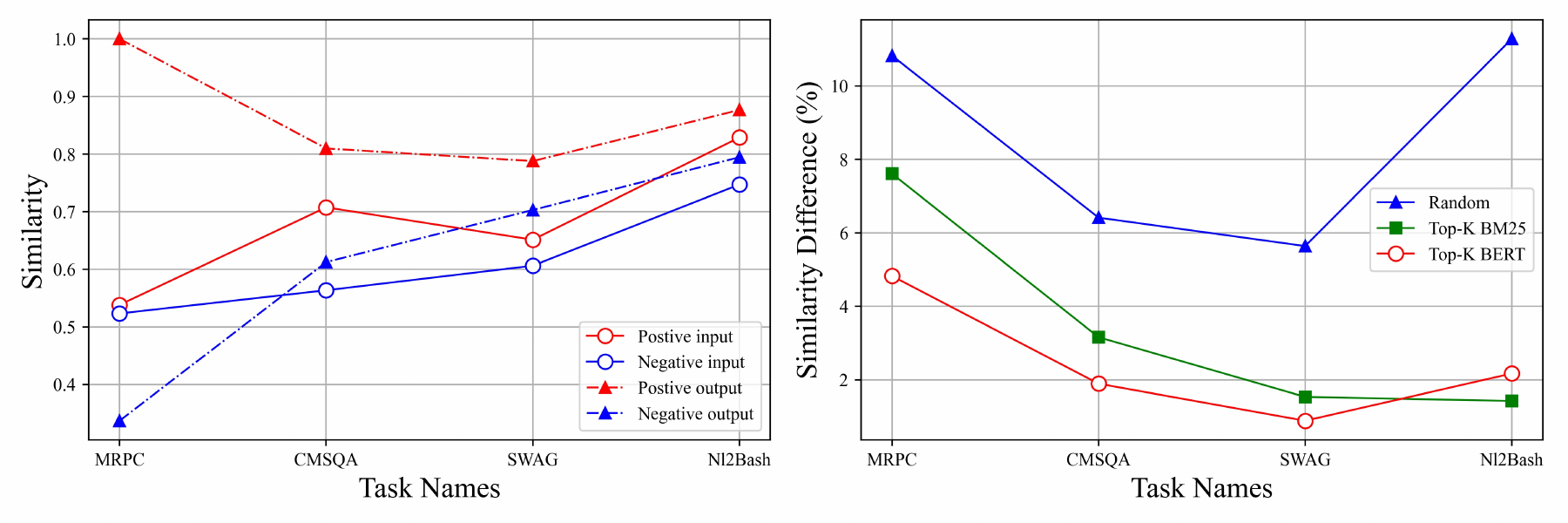}
        % \vspace{-6ex}
    \caption{\textbf{Left}: Comparison of similarity between the input/output of positive and negative demonstration examples and the input/output of the test case across four tasks for EPR. \textbf{Right}: Difference in average similarity between the output of test case and retrieved exemplars for EPR and each of the three learning-free prompt retrieval methods.  We use Llama3 (8B) as the main LLM.}
 
 \label{fig:3:llma3}
\end{figure}

\begin{table}[h]
\centering
\caption{Main results of MLSM and TTF on four datasets when using Llama 3 (8B) as the main LLM. Llama 3 (8B) performs worse on NL2Bash because of the repetitive generation}
\label{tab:mainllama3}
\begin{adjustbox}{max width=\linewidth, center}
\begin{tabular}{l|llllll}
\hline
Method     & MRPC  & CMSQA & SWAG  & NL2B. & Avg.  & Avg. (w/o NL2B.) \\
\hline
Random     & 67.65 & 68.39 & 74.67 & 9.89  & 55.15 & 70.23            \\
Top-K BERT & 72.28 & 68.29 & 74.03 & 9.77  & 56.09 & 71.54            \\
MLSM       & 71.32 & 68.88 & 76.69 & 15.67 & 58.13 & \textbf{72.30}            \\
\cline{2-7}
EPR        & 72.30 & 66.77 & 74.14 & 9.87  & 55.77 & 71.07            \\
TTF        & 72.79 & 68.80 & 76.77 & 12.37 & 57.67 & \textbf{72.79}           \\
\hline
\end{tabular}
\end{adjustbox}
\end{table}

\noindent\textbf{Generalization of of \HOne and \HTwo}: In the main body of our paper, we validate \HOne~and \HTwo~using GPT-Neo and GPT-2 XL. To extend this to more advanced LLMs, we utilize Llama 3 (8B) for the learning-based method EPR on four datasets (i.e., MRPC, CMSQA, SWAG, and Nl2Bash) due to the high cost of data collection for EPR's proxy task. As illustrated in Fig. \ref{fig:2:llma3} (Left), different tasks exhibit distinct preferences for specific layers, and the CKA distribution across various tasks shows significant diversity among different pre-trained layers of BERT in Fig. \ref{fig:2:llma3} (Right). These results support \HOne, indicating that learning-based methods can effectively aggregate multi-level (layer) linguistic similarities across tasks. Additionally, as depicted in Fig. \ref{fig:3:llma3} (Left), positive exemplars have consistently higher input-output similarities with test cases than negative ones. Furthermore, as shown in Fig. \ref{fig:3:llma3} (Right), the exemplars chosen by EPR have outputs more similar to the test case than those selected by unsupervised competitors, supporting \HTwo. Thus, our findings can be generalized to more advanced LLMs.

\noindent\textbf{Generalization of MLSM and TTF}: We further verify the generalization capabilities of \textbf{MLSM} and \textbf{TTF} to more advanced LLMs using Llama 3 (8B) and GPT-3.5 on four tasks. First, we compared these methods against both supervised (EPR) and unsupervised approaches (Top-K BERT, Random) with Llama 3 (8B), as shown in Table \ref{tab:mainllama3}. The results demonstrate that MLSM consistently surpasses Top-K BERT, while TTF achieves the highest performance overall. We also evaluated MLSM and TTF against Top-K BERT on Llama 3 (8B) and GPT-3.5 across varying shot numbers in Table \ref{tab:transfergptllama}. Both methods generally outperform Top-K BERT, except for TTF when using 20 shots. We speculate that GPT-3.5 may learn incorrect patterns from selected exemplars when it could answer correctly using its inherent knowledge, but the implicit prediction by TTF and EPR is wrong. Moreover, advanced LLMs are more sensitive to the instruction prompt choice rather than exemplar, particularly when the shot of exemplars reaches a certain threshold (e.g., 3 shots) \cite{chen2023robust, yuan2024focused}. In summary, our methods demonstrate strong generalization across advanced LLMs.

\section{Connection with Explanatory Work of ICL}
\label{app:icl}
Our work presents two hypotheses regarding the types of similarity measurements acquired by learning-based demonstration selection methods: Integrating task-agnostic similarities of different levels between the input of exemplars and test cases (\HOne), Incorporating task-specific similarity between the output of exemplars and test cases (\HTwo). While we have quantitatively validated both hypotheses in Sec. \ref{sec:assumption},  we qualitatively support both hypotheses by demonstrating the exemplars selected based on the corresponding similarity measurements will contribute to the ICL performance based on the explanatory mechanisms of ICL.

\noindent\textbf{Qualitative  Validation of \HOne}: \HOne~argues learning-based exemplar selection methods retrieve exemplars with multi-level analogs to the test case. These exemplars are more likely to lead LLMs to correct predictions than dissimilar ones when they contain relevant patterns (i.e., token and token sequences that aid correct predictions) for the test case. For example, previous investigation \cite{iclinductionhead, indcutionhead2} proposed a possible inner working of ICL that LLMs can learn from surface patterns in the demonstration sequence, such as copying tokens from contextual prompts. Furthermore, recent research \cite{Repetitionsiclr24} empirically demonstrated that increased contextual co-occurrences will strengthen the connection between two tokens during generation caused by the maximizing likelihood objective of LLMs. These insights suggest that influential demonstration exemplars may exhibit more token or phrase-level correspondence with the test case corresponding to the low-level similarities in the lower or middle layers of pre-trained BERT, significantly influencing LLM outputs and supporting \HOne.

\noindent\textbf{Qualitative  Validation of \HTwo}: In line with the qualitative validation of \HOne, we illustrate the exemplar with similar input and output to test cases also contributes to the performance of ICL. Prior work has demonstrated that ICL typically learns input-output relation from exemplars even for a genuinely novel task the LLM cannot know from pre-training~\cite{randomlabeliswrong1,halawi2023overthinking, zhao2024unveiling}. Moreover, \citet{randomlabeliswrong1} further proposed that LLMs prefer utilizing information closer to the query rather than treating all available information equally. Hence, if the exemplar selection method successfully learns the output similarity via the proxy task, it selects demonstration examples exhibiting useful input-output correlations for the test case due to their shared relevant input-output correlations and positions it closely to the test query in the prompt. These advantages align with the previously mentioned underlying working mechanisms of LLMs, thereby validating \HTwo.
\begin{table}[h]
\centering
\caption{Results of cross-LLM Transferability Validation of TTF and MLSM on Llama 3 8B and GPT 3.5 Turbo.}
\label{tab:transfergptllama}
\begin{adjustbox}{max width=\linewidth, center}
\begin{tabular}{cclrrrrr}
\hline
\multicolumn{8}{c}{TTF   Verification}    \\
\hline
\multicolumn{1}{l}{LLM}   & Shot                                        & \multicolumn{1}{c}{Method} & \multicolumn{1}{l}{SST-5} & \multicolumn{1}{l}{MRPC}  & \multicolumn{1}{l}{QNLI}  & \multicolumn{1}{l}{CMSQA}   & \multicolumn{1}{l}{Avg.} \\
                          &                                            & Top-K BERT                 & 48.50                     & 70.34                     & 78.13                     & 63.05                       & 65.00                    \\
                          & \multirow{-2}{*}{3}                        & TTF                        & 48.68                     & 70.83                     & 77.96                     & 63.55                       & \textbf{65.26}                    \\
                          \cline{3-8}
                          &                                            & Top-K BERT                 & 49.41                     & 71.32                     & 77.25                     & 60.52                       & \textbf{64.63}                    \\
                          &                                            & TTF                        & 47.96                     & 66.91                     & 77.34                     & 60.94                       & 63.29                    \\
\multirow{-5}{*}{GPT 3.5} & \multirow{-3}{*}{20}                       & EPR                        & 49.14                     & 64.22                     & 77.03                     & 59.79                       & 62.55                    \\
 \cline{3-8}
                          &                                            & Top-K BERT                 & \multicolumn{1}{l}{72.28} & \multicolumn{1}{l}{71.28} & 73.73                     & 68.29                       & 71.40                    \\
\multirow{-2}{*}{Llama 3} & \multirow{-2}{*}{20}                       & TTF                        & 72.79                     & 72.79                     & 77.72                     & 68.80                       & \textbf{73.03}                    \\
\hline
\multicolumn{8}{c}{MLSM Verification}                                                                                                                                                                                                            \\
\hline
\multicolumn{1}{l}{LLM}   & Shot                                       & \multicolumn{1}{c}{Method} & \multicolumn{1}{l}{SST-5} & \multicolumn{1}{l}{MRPC}  & \multicolumn{1}{l}{GeoQ.} & \multicolumn{1}{l}{NL2Bash} & \multicolumn{1}{l}{Avg.} \\
                          &             & Top-K BERT                 & 49.50                     & 70.34                     & 17.14                     & 63.52                       & 50.13                    \\
                          & \multirow{-2}{*}{{3}} & MLSM                       & 49.32                     & 70.59                     & 18.00                     & 64.56                       & \textbf{50.62}                    \\
                          \cline{3-8}
                          &                                            & Top-K BERT                 & 49.41                     & 71.32                     & 4.64                      & 60.52                       & 46.47                    \\
\multirow{-4}{*}{GPT 3.5} & \multirow{-2}{*}{20}                       & MLSM                       & 50.23                     & 74.02                     & 5.36                      & 68.24                       & \textbf{49.46}                    \\
  \cline{3-8}
                          &                                            & Top-K BERT                 & 72.28                     & 71.28                     & 0.00                      & 9.77                        & 38.33                    \\
\multirow{-2}{*}{Llama 3} & \multirow{-2}{*}{20}                       & MLSM                       & 72.79                     & 72.79                     & 0.00                      & 15.67                       & \textbf{40.31}             \\   
\hline
\end{tabular}
\end{adjustbox}
\end{table}

\begin{table}[htbp]
\centering
\caption{Main results of MLSM and TTF when using GPT Neo as the main LLM. $\dagger$ represents the probability that the performance of MLSM exceeds that of Top-K BERT is over $95\%$ by t-test. Org represents the performance reported in our original version. }
\label{tab:ttest}
\begin{adjustbox}{max width=\linewidth, center}
\begin{tabular}{l|llllll}
    \hline
\multicolumn{7}{c}{Main results on the classification task.}      \\
    \hline
Method     & SST-5                     & MRPC                   & QNLI                    & CMSQA                  & SWAG                   & AVG.                   \\
  \hline
Top-K BERT & $32.64$                   & $69.70$                & $61.94$                 & $35.25$                & $41.46$                & $48.20$                \\
MLSM (Org) & $33.15$                   & $69.87$                & $65.02$                 & $37.26$                & $41.49$                & $49.36$                \\
MLSM       & ${35.00\pm 1.77}^\dagger$ & $69.69\pm0.29$         & $65.10\pm0.11^\dagger$  & $38.07\pm0.81^\dagger$ & $41.82\pm0.32$         & $\mathbf{49.94}\pm0.60^\dagger$ \\
\cline{2-7}
EPR (Org)  & $36.88$                   & $81.37 $               & $77.87$                 & $38.74$                & $43.39$                & $55.65$                \\
TTF        & $42.04\pm1.50$            & $74.18\pm0.58$         & $85.15\pm1.00$          & $46.39\pm1.55$         & $56.51\pm{0.69}$       & $\mathbf{60.85}\pm{0.27}$       \\
    \hline
\multicolumn{7}{c}{Main results on   the generation  task.}                                          \\
    \hline
Method     & WebQs                     & GeoQ.                  & NL2B.                   & MTOP                   & SMCal.                 & AVG.                   \\
  \hline
Top-K BERT & $14.13$                   & $64.44$                & $53.15$                 & $51.49$                & $44.76$                & $45.59$                \\
MLSM (Org) & $16.14$                   & $68.93$                & $56.11$                 & $54.05$                & $47.72$                & $48.59$                \\
MLSM       & ${15.65\pm 0.47}^\dagger$ & $69.14\pm0.19^\dagger$ & $56.24\pm 1.27^\dagger$ & $53.92\pm0.20^\dagger$ & $47.59\pm0.19^\dagger$ & $48.51\pm0.17^\dagger$ \\
\hline
\end{tabular}
\end{adjustbox}
\end{table}
\section{Statistical Significance}
\label{app:Significance}
To strengthen our evaluation, we re-ran MLSM and TTF for the main experiments in Table \ref{man:cls} and Table \ref{man:gen} using GPT-Neo as the LLM. We report the average accuracy and standard deviation for both methods and statistical significance for the comparison between MLSM and Top-K BERT in Table \ref{tab:ttest}. The results demonstrate the effectiveness of both methods, particularly MLSM, which shows stable performance improvements, which may be attributed to the used loss function.

\section{Theoretical Foundation Of MLSM and TTF}
\label{app:foundation}
While MLSM and TTF are naturally supported by our findings (\HOne and \HTwo) as they are two implementations of these findings, we provide a preliminary theoretical foundation for both methods. Specifically, MLSM treats different layers as experts and uses the loss function $\mathcal{L} = - \sum_{i=1}^{n_l}\hat{\mathbf{e}} \cdot \mathbf{e}_i$ to ensemble them for demonstration selection. This approach is theoretically proportional to mutual information $I(E, \hat{E})$ and inversely proportional to selection entropy $H(\hat{E})$, maximizing expert agreement and ensuring stable selection. The detailed proof is available in \cite{DBLP:conf/nips/ZhangHHF22}. For TTF, we conduct preliminary theoretical analysis showing how features from layers before the final classification task heads can model input-output distribution in Section \ref{sec:method}.

\begin{figure}[t]
\includegraphics[width=\linewidth]{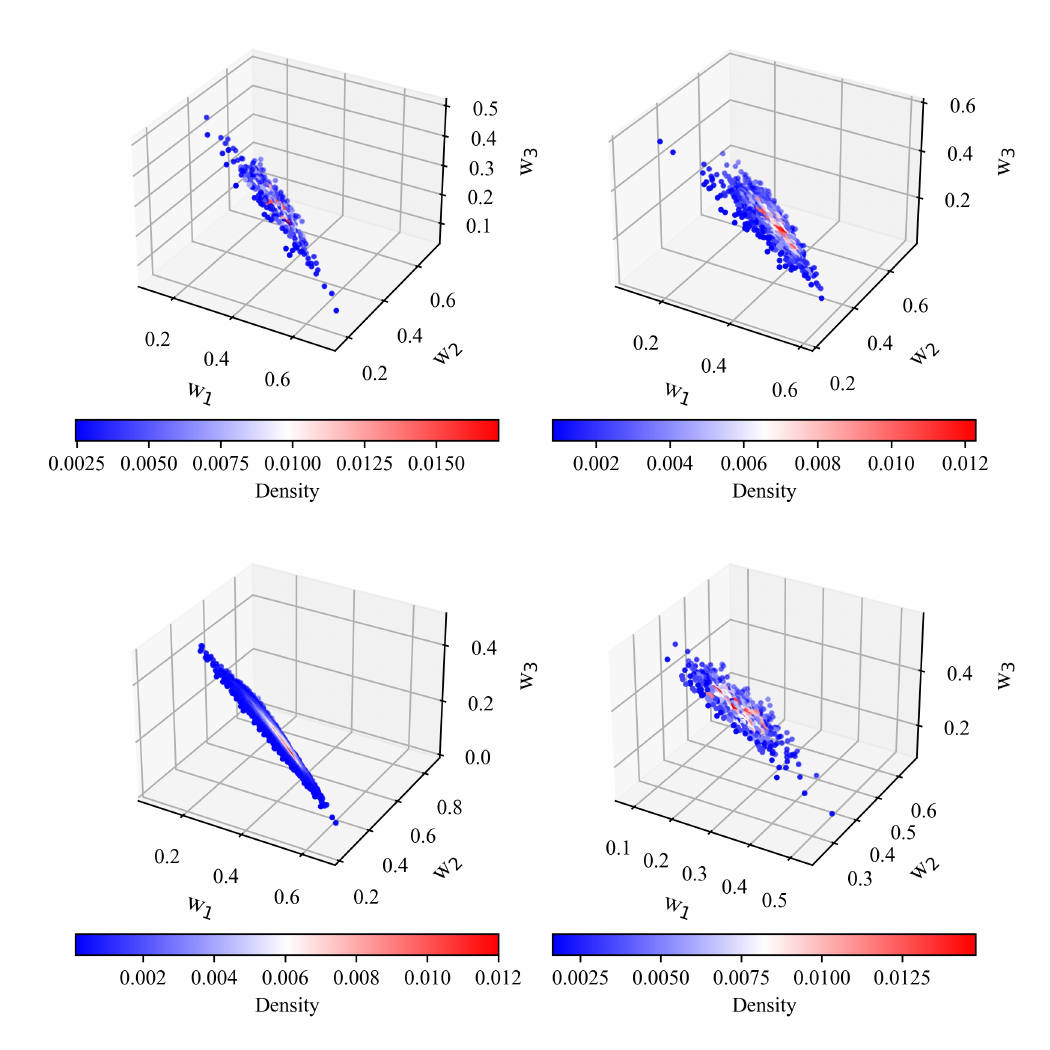}
\caption{Probability density distribution of aggregated weights for $n_l$ layers of MLSM, with $n_l = 3$ for MRPC, CMSQA, SWAG, and Nl2Bash, presented from top-left to bottom-right. The weights $w_1$, $w_2$, $w_3$ correspond to layers from low to high. The mean (standard deviation) of the weights are as follows: 0.35 (0.08), 0.43 (0.08), 0.21 (0.07) for MRPC,  0.34 (0.07), 0.48 (0.08), 0.17 (0.06) for CMSQA,  0.34 (0.07), 0.50 (0.08), 0.15 (0.06) for SWAG and  0.26 (0.06), 0.43 (0.07), 0.30 (0.07) for Nl2Bash.}
  \centering
\label{fig:5aggreweigh}
\end{figure}
\section{Analysis of aggregation weight}
\label{app:weightanalysis}
We analyze the probability density distribution of aggregation weights of MLSM on four datasets in Fig. \ref{fig:5aggreweigh}. The results show that: 1) Different datasets exhibit varying weight probability density distributions and mean values, indicating that MLSM adaptively adjusts the weights of each layer to maximize agreement for demonstration retrieval. 2) The weights $w_1$ and $w_2$ are often higher than $w_3$, suggesting that MLSM focuses more on lower-level features, possibly due to the greater similarity of features extracted from these layers. Although MLSM's performance is impressive, this method is just one possible instance of our proposed \HOne. Other alternatives, such as integrating MLSM with training examples from the proxy task of learning-based methods, may also be viable, which we leave in future exploration.

\section{Running Efficiency}
\label{app:compcost}
Take the experiments on the QNLI dataset using a V100 GPU as an example. QNLI, a natural language inference task, comprises 5,463 test samples and 104,707 demonstration samples. For MLSM, in an online streaming scenario with a batch size of 1 (i.e., only one test point is observed during inference), this method processes approximately 1.6–1.7 data points per second. However, as indicated in Ablation of Batchsize for MLSM in Sec. \ref{sec:5}, MLSM benefits significantly from larger batch sizes. In this case, with batch sizes of 8 and 64, MLSM can process approximately 4 and 32 data points per second, respectively. Additionally, the GPU memory overhead for MLSM is small (400–800 MB), enabling multi-process execution to accommodate deployment requirements. In comparison, TTF can process approximately 60 data points per second.